\PassOptionsToPackage{sort&compress}{natbib}
\documentclass[preprint,12pt,3p]{elsarticle}

\usepackage{multirow}

\usepackage{amssymb}
\usepackage{amsmath}
\usepackage{comment}

\usepackage{algorithm}
\usepackage[noend]{algpseudocode}
\usepackage{url}
\usepackage{booktabs}
\usepackage{pifont}

\usepackage{graphicx}
\usepackage{caption}
\usepackage{subcaption}

\usepackage[table,xcdraw]{xcolor}

\usepackage{makecell}
\newcommand{\cell}[2]{%
\makecell[c]{\footnotesize   #1 \\ {\scriptsize #2}}%
}

\journal{Journal of Computers in Biology and Medicine}

\begin{document}

\begin{frontmatter}



\title{xMICD: Explainable Representation\\ of Multiple ICD Codes}

\author[label1,label2,label4]{Pat Vatiwutipong} 

\author[label3]{Kumkup Keeratisiwakul} 

\author[label3]{Albert Phuoc Kien Van Truong} 

\author[label6]{Nutcha Yodrabum} 

\author[label5]{Wasin Pansiritanachot} 

\author[label2,label4]{Marvin N. Wright} 

\author[label1]{Thanapon Noraset\corref{cor1}} 
\ead{thanapon.nor@mahidol.ac.th}
\cortext[cor1]{Corresponding author}

\affiliation[label1]{organization={Faculty of Information and Communication Technology},
            addressline={Mahidol University}, 
            city={Nakhon Pathom},
            country={Thailand}}

\affiliation[label2]{organization={Faculty of Mathematics and Computer Science},
            addressline={University of Bremen}, 
            city={Bremen}, 
            country={Germany}}

\affiliation[label4]{organization={Leibniz Institute for Prevention Research and Epidemiology - BIPS},
            city={Bremen},
            country={Germany}}
            
\affiliation[label3]{organization={Department of Transdisciplinary Science and Engineering, School of Environment and Society},
            addressline={Institute of Science Tokyo}, 
            city={Tokyo},
            country={Japan}}

\affiliation[label5]{organization={Department of Emergency Medicine, Faculty of Medicine Siriraj Hospital, Mahidol University},
            city={Bangkok},
            country={Thailand}}
            
\affiliation[label6]{organization={Department of Surgery, Faculty of Medicine Siriraj Hospital, Mahidol University},
            city={Bangkok},
            country={Thailand}}


\begin{abstract}

\noindent \textbf{Background: } Electronic Health Records (EHRs) are widely used for clinical risk prediction using machine learning. International Classification of Diseases (ICD) codes provide structured information about patient diagnoses, but representing them effectively remains challenging. Existing approaches often face a trade-off between predictive performance and interpretability: grouping-based representations are interpretable but may lose information, while embedding-based representations achieve strong predictive performance but are difficult to interpret.

\noindent\textbf{Methods: } We propose Explainable Representation of Multiple ICD Codes (xMICD), a method for constructing low-dimensional patient representations from sets of ICD codes. xMICD combines clinically meaningful diagnostic groupings with similarity in a pre-trained ICD embedding space. Instead of using binary group membership, the method assigns codes to groups via similarity-based relative assignments, yielding features that reflect how closely a patient’s diagnoses align with each clinical group.

\noindent\textbf{Results: } Experiments on large-scale EHR datasets demonstrate that xMICD achieves predictive performance comparable to embedding-based representations such as ICD2Vec across multiple clinical prediction tasks. At the same time, the resulting features remain clinically interpretable because each dimension corresponds to a recognizable diagnostic group. xMICD therefore provides a practical way to integrate embedding-based semantic relationships into interpretable clinical feature spaces for machine learning models.

\end{abstract}



\begin{keyword}
International Classification of Diseases codes\sep Electronic Health Records\sep Medical code embeddings\sep Interpretable features\sep Explainable AI



\end{keyword}
\end{frontmatter}


\section{Introduction}
\label{sec1}

Electronic Health Records (EHRs) are widely used to develop machine learning (ML) models for clinical risk prediction and decision support. Among the many elements recorded in EHRs, International Classification of Diseases (ICD) codes play a key role, as they summarize patients’ diagnoses in a standardized format \cite{Menachemi2011, WHO1978, Quan2008}. However, using ICD codes directly in ML models is not straightforward. The large number of distinct codes results in high-dimensional and sparse feature spaces when one-hot encoded, which can increase model complexity and reduce generalization performance.

Existing methods for representing ICD codes generally follow two directions. The first uses clinically defined groupings, such as comorbidity indices or ICD chapters and blocks. These representations are compact and easy to understand, but they often lose detailed information and may not achieve strong predictive performance \cite{Atutxa2017, Xu2019}. The second direction relies on deep learning–based embeddings, which learn dense vector representations that capture relationships between codes. These approaches often perform well in prediction tasks, but the learned dimensions are abstract and difficult to interpret in clinical terms \cite{Deimazar2023, Zikos2021}. This creates a trade-off between interpretability and predictive accuracy.

Interpretability is especially important in healthcare, where ML models may influence high-stakes decisions. It is not enough for a model to be interpretable at the algorithm level; the input features should also be understandable to clinicians and other stakeholders \cite{zytek2022, Senel2020}. Even simple models can be difficult to explain if their input features lack clear clinical meaning. Therefore, designing ICD representations that are both informative and interpretable is an important step toward trustworthy clinical ML systems.

In this work, we propose xMICD (Explainable Representation of Multiple ICD Codes), a method that aims to balance predictive performance and interpretability. xMICD represents a patient’s set of ICD codes as a single, low-dimensional vector. Each dimension corresponds to a clinically meaningful group of codes (e.g., ICD-10-CM blocks).

Instead of using only binary indicators for group membership, xMICD employs a relative assignment mechanism based on similarity in a pre-trained ICD embedding space. Specifically, the value of each dimension reflects the similarity between a patient’s diagnoses and a predefined clinical reference group, allowing each dimension to capture graded relatedness rather than simple presence or absence.

Conceptually, this representation is inspired by anchor-based interpretable embeddings, where each dimension reflects similarity to predefined reference anchors in the representation space \cite{opitz2025interpretable,wang2025ldir}. By anchoring each dimension to clinically defined diagnostic groups, xMICD produces a compact and interpretable relative representation of ICD codes.

Our contributions are as follows:
\begin{itemize}
    \item We introduce xMICD, a method that converts multiple ICD codes into a low-dimensional, clinically structured feature vector for standard ML models.
    \item We integrate pre-trained ICD embeddings with clinically defined groupings to retain semantic relationships while preserving feature-level interpretability.
    \item We evaluate xMICD on a large-scale EHR dataset against common ICD representations across predictive performance in several downstream tasks, similarity preservation, and interpretability. xMICD achieves performance comparable to embedding-based methods while retaining clinically grounded structure absent in dense black-box embeddings.
\end{itemize}


\section{Background}

\subsection{Challenges of Using ICD Codes With Machine Learning}

The International Classification of Diseases (ICD) was first introduced in 1900 as a system for recording causes of death. Since then, the World Health Organization (WHO) has revised it multiple times, leading to ICD-9 and ICD-10, which are still widely used today \cite{WHO1978, Quan2008}. ICD-10 contains more than 12,000 codes, and extended versions such as ICD-10-CM include around 70,000 codes \cite{WHO2016, cdc2022}. 

ICD-10-CM codes follow a structured format. The first three characters represent a broad disease category. Codes are grouped into chapters based on the first letter, and further divided into blocks within each chapter. Additional characters provide more detailed information, such as anatomical site, severity, or type of encounter. For example, in the code S52.001A, “S” indicates injury-related conditions, “52” refers to a fracture of the forearm, and the remaining characters specify further details, including side and encounter type. ICD-10 codes have a similar structure but are generally less detailed than ICD-10-CM \cite{WHO2016, cdc2022}.

While ICD codes are clinically meaningful, they create practical challenges for machine learning. When represented as one-hot vectors, the large number of distinct codes leads to very high-dimensional and sparse feature spaces. This can increase computational cost and make it harder for models to generalize well \cite{Bishop2006, Xu2019}. For this reason, many studies have proposed transforming ICD codes into lower-dimensional representations. These approaches are reviewed in Section~\ref{sec:relate}.

\subsection{Interpretable Features}
\label{GoodFeatures}

In machine learning, interpretability generally refers to how easily humans can understand how a model works, while explainability focuses on how well the reasons behind a model’s predictions can be communicated \cite{rudin2019stop,Longo2024, Hakkoum2022}. In healthcare, limited interpretability may reduce trust, slow adoption, and raise ethical or legal concerns \cite{Vellido2019, Zardai2023}. Clinicians who cannot clearly understand how a model reaches its conclusions may hesitate to rely on it in practice.

Importantly, interpretability is not only about the model itself but also about the features it uses. Even simple models, such as linear regression, can be difficult to interpret if their input features lack clear meaning \cite{Longo2024}. Therefore, in clinical ML applications, features should be understandable and clinically relevant to domain experts \cite{zytek2022}.

According to \citet{zytek2022}, feature properties can be grouped into three categories. The first includes properties needed for effective modeling, such as predictive strength and compatibility with the chosen algorithm. The second includes properties related to interpretability, such as readability, clear descriptions, and alignment with real-world clinical concepts. The third includes general properties relevant to all stakeholders, such as meaningfulness, traceability back to the original data, and the ability to reason about how changes in inputs affect outputs \cite{Preece2018}. 

These considerations show that choosing an appropriate feature representation is as important as selecting the learning algorithm itself. This motivates our development of xMICD, which aims to produce features that are both model-ready and clinically interpretable.


\section{Related Work}
\label{sec:relate}

Existing approaches for representing ICD codes in machine learning fall into two main categories: grouping-based binary representations and embedding-based representations.

\subsection{Grouping-Based Binary Representations}
Grouping-based representations reduce the dimensionality of ICD codes by mapping them into clinically defined categories and representing each category using binary indicators.

The ICD coding system itself defines a hierarchical structure consisting of chapters and blocks, which group related diagnoses within the taxonomy. These hierarchical levels can be encoded as binary features indicating whether a visit contains at least one diagnosis code within each group \cite{cartwright2013icd,dugan2017international}.

Beyond the native ICD hierarchy, several derived grouping systems have been developed. The Clinical Classifications Software (CCS) aggregates ICD-9-CM codes into mutually exclusive clinical categories \cite{CCS2017}. For ICD-10-CM, the related Clinical Classifications Software Refined (CCSR) provides a similar mapping that organizes diagnosis codes into clinically meaningful groups \cite{CCSR2020}.

Another widely used approach focuses specifically on comorbidity measurement. The Charlson Comorbidity Index (CCI) summarizes diagnoses into 17 predefined comorbidity groups \cite{charlson1987}. The Elixhauser Comorbidity Index (ECI) later expanded this framework to 31 comorbidity categories derived from ICD-9-CM codes, where each category is typically encoded as a binary indicator \cite{elixhauser1998}. These indices were originally designed for severity adjustment and outcome comparison in clinical and epidemiological studies \cite{ludvigsson2021adaptation,bannay2016best,sharma2021comparing}.

These representations are easy to interpret and align with clinical reasoning. However, by reducing each group to a binary indicator, they treat heterogeneous diseases within the same category as equivalent. This simplification may obscure differences in severity or subtype and can limit predictive performance in some tasks \cite{Kansal2021,askar2024using,chervu2024development}.

\subsection{Deep Learning-Based Embedding Methods}
\label{sec-related-deep}

Embedding-based methods aim to learn dense vector representations for ICD codes that capture semantic relationships beyond manual groupings. Med2Vec is one of the early approaches, learning code embeddings from patient visit data \cite{choi2016}. ICD2Vec further leverages textual descriptions of ICD codes to generate embeddings, and patient-level representations are typically formed by averaging code vectors within a visit \cite{lee2023}.

More recent work incorporates temporal context. Models such as BEHRT and Med-BERT adapt Transformer architectures to represent sequences of visits, capturing disease progression and longitudinal dependencies \cite{li2020behrt,rasmy2021medbert}. In these approaches, embeddings become context-dependent rather than static.

Graph-based methods provide another direction. For example, GraphSAGE-MC models ICD codes as nodes in a graph defined by hierarchical relationships and introduces multi-code nodes to represent combinations of diagnoses \cite{Lui2025}. This design allows the model to represent unseen code combinations without retraining.

Embedding-based methods generally achieve strong predictive performance because they capture complex relationships between codes. However, their learned dimensions are abstract and do not correspond directly to clinically meaningful categories. This limits their transparency at the feature level, especially in high-stakes clinical settings.


\section{xMICD: The Proposed Method}
\label{sec:xMICD}

A common way to represent multiple ICD codes is to rely on clinical groupings.  
Let a visit be associated with ICD codes $
\{C_1, C_2, \ldots, C_n\}$. Assume that $m$ clinically meaningful groups (for example ICD Blocks or comorbidity categories) are available. A standard grouping-based representation defines, for each group $j$,
\[
G_j =
\begin{cases}
1 & \text{if any } C_i \text{ belongs to group } j, \\
0 & \text{otherwise.}
\end{cases}
\]
This representation is straightforward to interpret, as each dimension corresponds to a known clinical concept. However, it treats all codes within a group as equivalent and does not differentiate between, for example, mild and severe manifestations or between codes that are similar but assigned to different groups.

The xMICD framework relaxes this hard membership assumption. Instead of using binary group membership, it relies on similarity between ICD codes and a set of representative “anchors”. The resulting representation can still be read at the level of clinical groups, but each dimension reflects how strongly the codes in a visit relate to the corresponding anchor.

\subsection{xMICD Formulation}

Let $\{A_1, \ldots, A_m\}$ be a set of anchors. Anchors may correspond to individual ICD codes or to vectors in an ICD embedding space, as discussed further.

We first compute a raw similarity between each code $C_i$ and each anchor $A_j$,
\[
s_{ij} = \mathrm{Sim}(C_i, A_j),
\quad i = 1,\ldots,n,\; j = 1,\ldots,m,
\]
where $\mathrm{Sim}(\cdot,\cdot)$ is a chosen code–anchor similarity measure. This yields an $n \times m$ similarity matrix.

To remove scale differences between codes and to focus on the relative importance of anchors for each code, we normalize similarities within each row. For a fixed code $C_i$, we apply min–max scaling across all anchors,
\[
s'_{ij}
=
\frac{s_{ij} - \min_{j} s_{ij}}
{\max_{j} s_{ij} - \min_{j} s_{ij}}.
\]
Under this transformation, the anchor with the highest similarity to $C_i$ is mapped to 1, the anchor with the lowest similarity is mapped to 0, and other anchors take values between 0 and 1 according to their relative distances. The normalized values $s'_{ij}$ describe, for each code, the relative strength of each anchor compared to the others, rather than absolute similarity in the embedding space. This emphasizes contextual dominance within each code.

We then aggregate across codes at the anchor level. For each anchor $A_j$, we define
\[
S_j = \max_{i=1,\ldots,n} s'_{ij}.
\]
The vector $S = (S_1, \ldots, S_m)$ is the xMICD representation of the set of ICD codes.

The quantity $S_j$ does not measure absolute geometric proximity between codes and anchor $A_j$. Instead, it captures whether $A_j$ is relatively dominant for at least one of the codes in the visit. If $S_j = 1$, then $A_j$ is the top-ranked anchor for at least one code. If $S_j < 1$, $A_j$ never attains the highest normalized similarity, and low values indicate that it is consistently outperformed by other anchors within all codes in the visit. In this way, xMICD encodes contextual dominance of anchors across the codes rather than global similarity.

\begin{figure}[t]
\centering
\includegraphics[width=\linewidth]{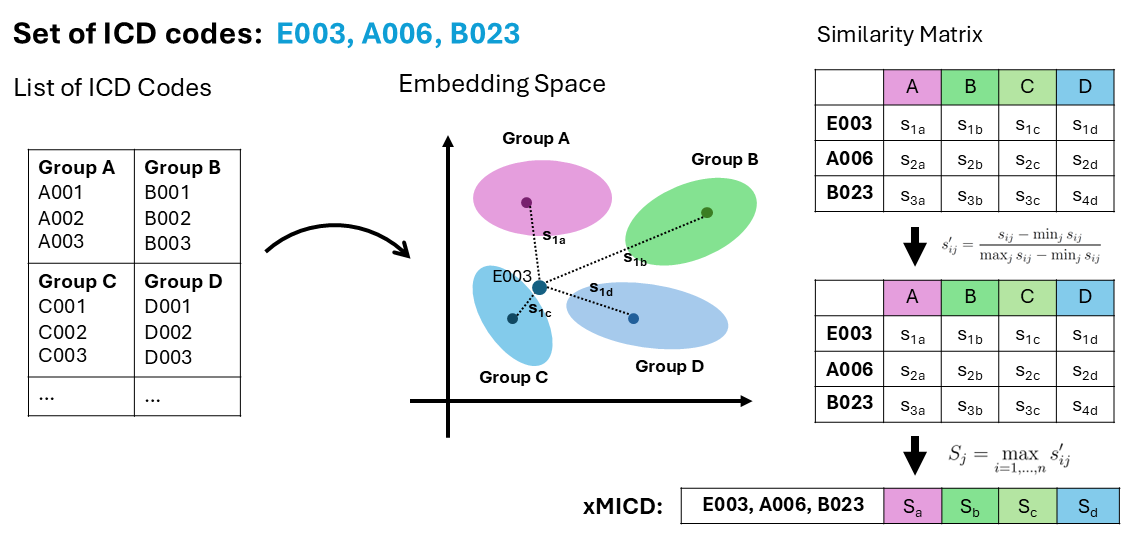}
\caption{Overview of the xMICD computation process. A set of ICD codes is mapped into an embedding space, similarities to anchor groups are computed, normalized within each code, and aggregated across codes to produce the final anchor-based representation.}
\label{fig:xmicd_pipeline}
\end{figure}

Figure~\ref{fig:xmicd_pipeline} illustrates the overall computation process of xMICD, from a set of ICD codes to the final anchor-based representation.

So far, two components remain to be specified: the similarity function $\mathrm{Sim}$ and the construction of the anchor set. We describe these choices next.

\subsection{Similarity Measures}

The similarity function $\mathrm{Sim}(\cdot,\cdot)$ can be defined at the code level or in an embedding space. We distinguish between these two perspectives.

\paragraph{Code similarity}

The first approach defines similarity directly between ICD codes. Such measures typically exploit structural or semantic information inherent in the ICD hierarchy. For example, ontology-based semantic similarity measures such as Wu--Palmer \cite{wu1994verb} and Leacock--Chodorow \cite{leacock1998combining} compute similarity based on the relative positions of concepts in the taxonomy.

These measures operate directly on the coding system and preserve relationships encoded in the ontology rather than relying on data-driven co-occurrence patterns. They have been used in prior work to compute semantic similarity between ICD codes when comparing patient diagnosis sets \cite{schneider2023improving}.

\paragraph{Embedding-based similarity}

The second approach defines similarity in a learned embedding space. Each ICD code is represented as a dense vector learned from clinical data, and similarity is computed using a standard vector similarity measure such as cosine similarity. Pre-trained models such as Med2Vec or ICD2Vec provide such embeddings (see Section~\ref{sec-related-deep}).\\

Unlike code-level measures, embedding-based similarity captures statistical relationships derived from data and may reflect co-occurrence or contextual patterns that are not explicit in the ICD hierarchy.

In this work, we primarily focus on embedding-based similarity, specifically ICD2Vec. This choice is motivated by the ability of embedding methods to capture richer relationships between diseases that may not be explicitly represented in the ICD hierarchy \cite{luo2024corelation,johnson2025clinvec}. In addition, using ICD2Vec as the similarity backbone allows us to directly compare the representations produced by xMICD with its underlying embedding space, providing a natural baseline for evaluating how the proposed method transforms and aggregates the original ICD2Vec representations.

\subsection{Anchor Selection}

Anchors act as reference points in the code space. We consider both clinically defined and data-driven constructions.

\paragraph{Clinically defined anchors}
Anchors may be derived from established clinical groupings such as CCI, ECI, or ICD Blocks. When a group contains multiple codes, a single representative can be selected as the most central element in the embedding space. For a group $k$, we define its anchor as
\[
A_k = 
\operatorname*{argmax}_{A \in \text{group } k}
\sum_{V \in \text{group } k}
\mathrm{Sim}(A,V),
\]
that is, the code whose embedding has the highest average similarity to other codes in the same group.

\paragraph{Data-driven anchors}
Anchors can also be defined directly from the embedding space without relying on predefined groupings. One option is farthest point sampling, which iteratively selects anchors that are maximally distant from those already chosen, thereby covering the space with diverse representatives. Another option is clustering, where a clustering algorithm is applied to the embeddings and cluster medoids are used as anchors. Such anchor selection strategies have been used in interpretable text embedding models such as LDIR \cite{wang2025ldir}.\\

The overall construction process combines a chosen similarity measure with either clinically defined or data-driven anchors. Through this design, xMICD provides a flexible framework for constructing clinically structured representations from ICD codes.

\subsection{Positioning of xMICD}

xMICD occupies an intermediate position between grouping-based binary representations and embedding-based dense representations. 

Compared to binary grouping methods, xMICD preserves graded semantic relationships through similarity-based relative assignment. Compared to dense embedding methods, xMICD constrains each dimension to correspond to a clinically interpretable anchor, improving feature-level transparency.

\begin{figure}[htbp]
    \centering
\includegraphics[width=0.7\linewidth]{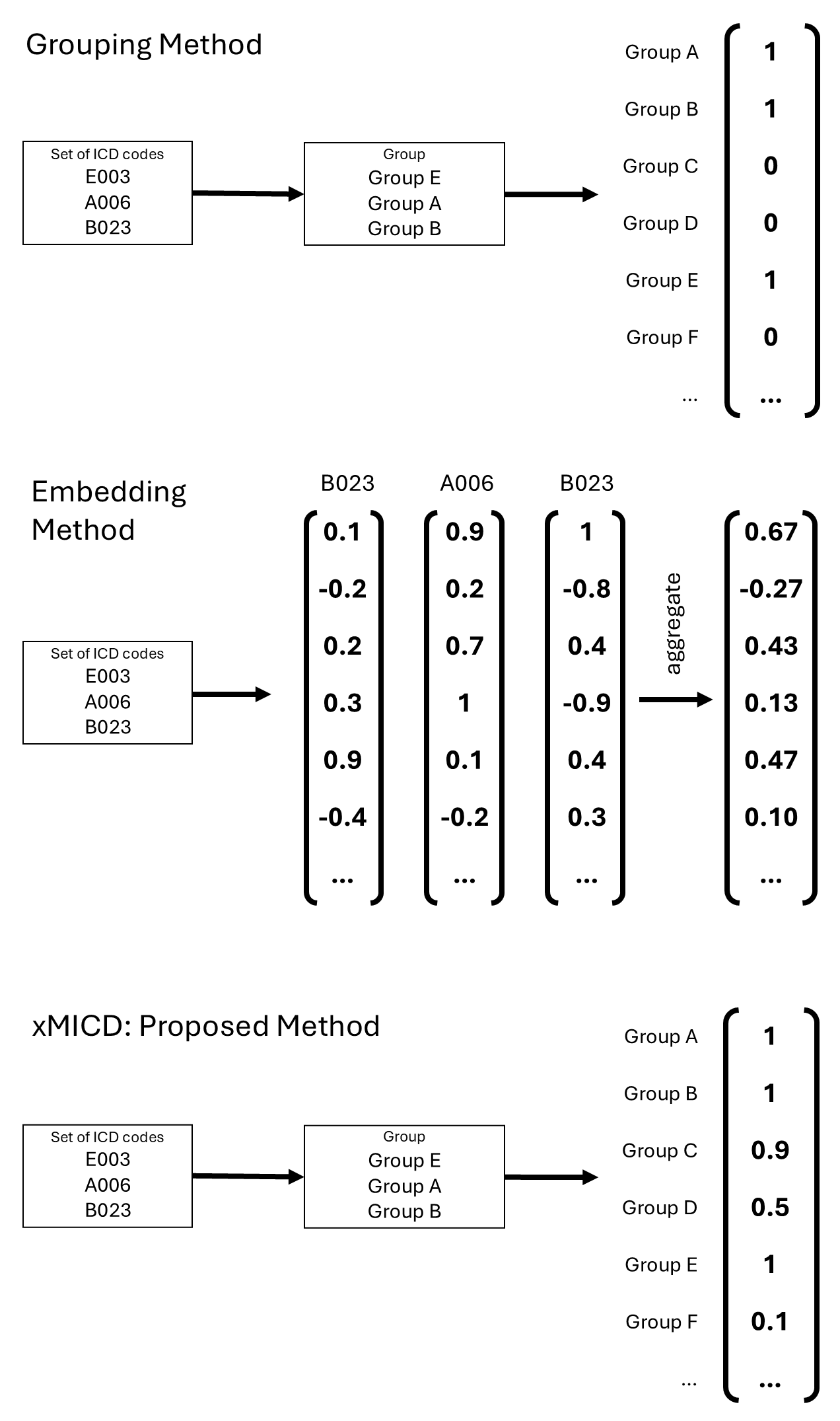}
    \caption{Positioning of xMICD between grouping-based and embedding-based representations for multi-ICD feature construction.}
    \label{fig:3methods}
\end{figure}

This positioning allows xMICD to balance semantic expressiveness and interpretability within a clinically structured representation space. Figure~\ref{fig:3methods} compares xMICD with grouping-based and embedding-based representations.

Note that xMICD values should be interpreted with care. Each dimension reflects the relative prominence of the corresponding anchor under the chosen similarity function, not a probability of belonging to that group and not a measure of disease severity. Because similarities are normalized across anchors for each ICD code and then aggregated across codes, a high xMICD value indicates that the anchor is relatively prominent for at least one diagnosis code in the visit.

\section{Experimental Setup}

\subsection{Setting}

\paragraph{Dataset and Preprocessing}

Experiments are conducted using two publicly available clinical datasets: MIMIC-IV-ED \cite{mimicived} and the eICU Collaborative Research Database (eICU-CRD) \cite{pollard2018eicu}. 

The MIMIC-IV-ED dataset contains 448,972 emergency department visits from 215,736 patients at Beth Israel Deaconess Medical Center between 2011 and 2019. Patients under 18 years of age or with missing triage acuity information were excluded, and outlier vital signs were removed. After preprocessing, 418,490 ED visits remained. The eICU-CRD is a multi-center intensive care dataset collected from hospitals across the United States. Records without ICD diagnosis codes or without discharge information were removed to ensure that both diagnostic features and outcome labels were available.

All ICD-9-CM codes were mapped to ICD-10-CM using the General Equivalence Mapping (GEM) to ensure consistency within a single coding system.

\paragraph{Baseline}

We compare several configurations of xMICD against established baselines, including raw ICD features, grouping-based binary representations, and embedding-based representations.

The raw ICD representation is constructed by one-hot encoding diagnosis codes appearing in the dataset. Although the ICD-10-CM system contains 24,354 diagnosis codes in total, only a subset appears in each dataset. In MIMIC-IV-ED, 10509 distinct codes occur in diagnoses recorded during the current ED visit and 14874 additional codes occur in inpatient admissions within the preceding five years, resulting in 25383 dimensions when combined. In contrast, the eICU-CRD dataset contains only 811 unique diagnosis codes.

Grouping-based baselines map ICD codes into predefined clinical groups and encode them as binary indicators, resulting in 31 dimensions for Elixhauser, 209 for ICD-10-CM Blocks, and 553 for CCSR. For embedding-based baselines, we use ICD2Vec embeddings with 1024 dimensions, where visit-level representations are obtained by averaging code embeddings.

For MIMIC-IV-ED, diagnoses from the current visit and the preceding five years are encoded separately and concatenated, doubling the dimensionality (62, 418, 1106, and 2048 respectively). In eICU, only diagnoses from the current ICU stay are used, so the dimensionality remains 31, 209, 553, and 1024.

\paragraph{xMICD configurations}
We construct xMICD under cosine similarity computed from ICD2Vec embeddings.
For anchors, we evaluate four strategies: medoids of ECI groups (31 anchors), medoids of ICD-10-CM Blocks (209 anchors), medoids of CCSR (553 anchors), and anchors selected using farthest point sampling (209 and 553 anchors).

\subsection{Evaluation}

We evaluate xMICD against other representations in terms of predictive performance on downstream tasks and similarity preservation.

\paragraph{Downstream tasks}

Representations of ICD codes are used as features for predictive tasks on both datasets.

For MIMIC-IV-ED, we follow the benchmark tasks introduced in \cite{Xie2022}: critical outcome prediction, hospitalization requirement, and revisit in 72h. ICD features are constructed using diagnoses from the current ED visit together with diagnoses recorded during inpatient admissions in the preceding five years to capture comorbidities. The dataset is randomly split into 80\% training and 20\% testing.

For eICU-CRD, we consider two commonly studied ICU outcomes: in-ICU mortality and prolonged ICU stay (defined as a length of stay greater than 3 days).  ICD features are derived only from diagnoses recorded during the corresponding ICU stay. The split is performed at the hospital level, assigning hospitals exclusively to either the training or test set, thereby providing external validation across institutions.

The experimental pipeline, predictive models, and evaluation procedures are kept consistent across both datasets. We evaluate two models, XGBoost and a multilayer perceptron (MLP) trained with Adam. Performance is measured using AUROC, and uncertainty is estimated via bootstrap resampling of the test set (200 replicates), following \cite{Xie2022}.

\paragraph{Similarity preservation}

In natural language processing, representation quality is often evaluated by comparing representation-induced similarity with a human-annotated ground-truth similarity. 
However, in the clinical domain, no such gold-standard similarity exists at the patient level.

We therefore evaluate similarity preservation by assessing the consistency of patient-level similarity structures across representations.

For each representation, patient-level similarity was computed using cosine similarity between visit-level feature vectors. We then computed Spearman’s rank correlation between the resulting similarity matrices across representations. This evaluates the consistency of patient similarity structures across different representations.

\subsection{Interpretability Assessment}
\label{sec:criteria}

Unlike predictive performance and similarity, feature-level interpretability does not have a widely accepted quantitative metric. Instead, prior work has proposed qualitative criteria that characterize whether features are actionable and clinically meaningful.

We therefore adopt the framework introduced in Section~\ref{GoodFeatures} to evaluate interpretability using a set of qualitative properties. 
Each property is assessed using three ordinal levels, indicating whether it is present, partially present, or absent for a given representation.
This allows us to systematically compare baseline representations and xMICD in terms of their interpretability characteristics.

The evaluated properties are: \textit{Readable:} Features are expressed in recognizable, domain-appropriate vocabulary rather than opaque or coded representations. \textit{Understandable:} Features refer to real-world quantities that users can reason about in context. \textit{Meaningful:} Features align with clinically plausible or domain-relevant concepts. \textit{Trackable:} Features have clear data lineage and can be traced back to the original input data. \textit{Simulatable:} The feature computation process can be reconstructed or reasoned about from available information. A detailed discussion of these criteria is provided in \cite{zytek2022}. 


\section{Results}

\subsection{Predictive performance on downstream tasks}

\begin{figure}[htbp]
\centering
\includegraphics[width=\linewidth]{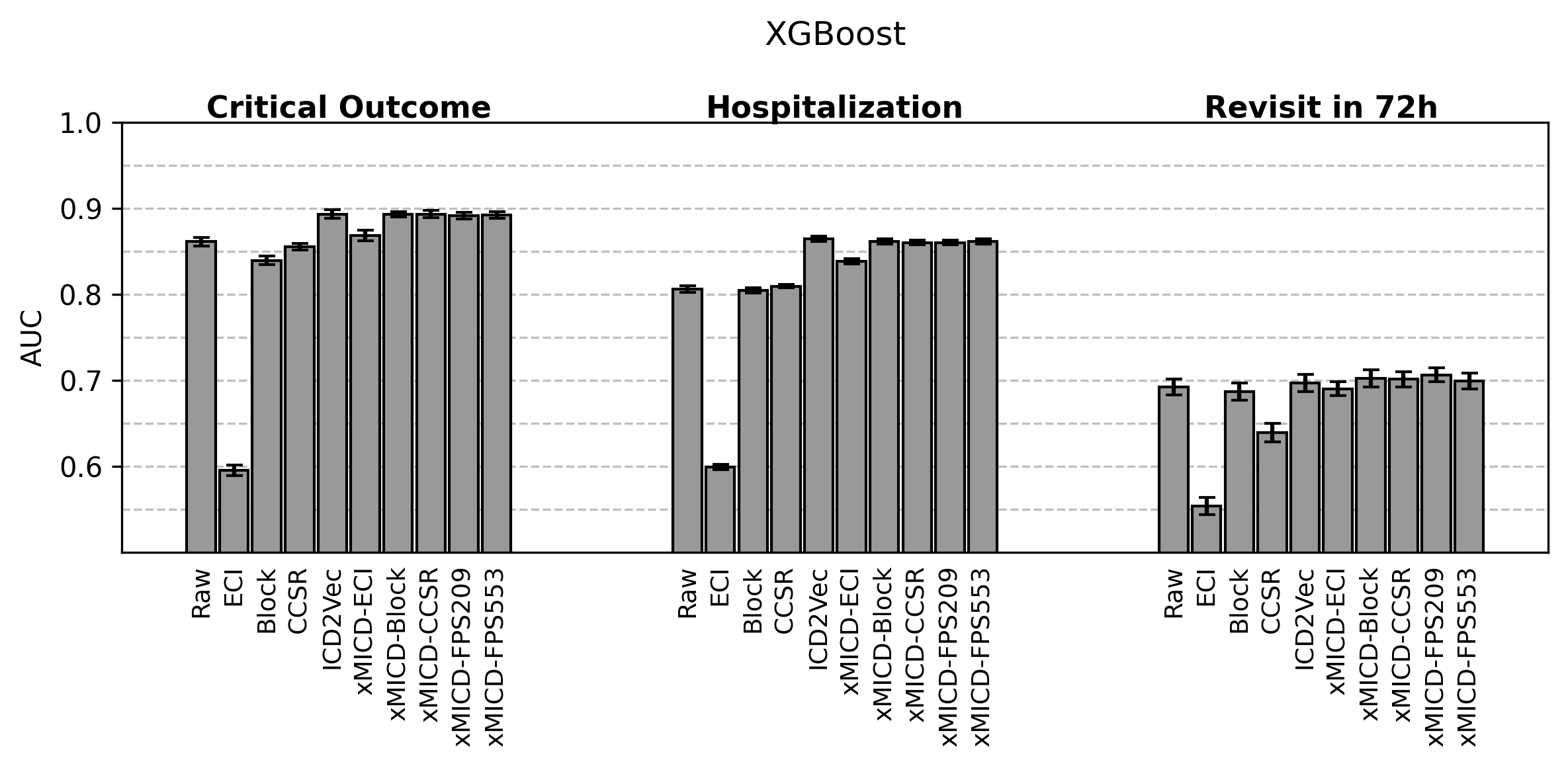}
\caption{Predictive performance of different ICD representations on the MIMIC-IV-ED dataset using XGBoost.}
\label{fig:xgboost_task123}
\end{figure}

\begin{figure}[htbp]
\centering
\includegraphics[width=\linewidth]{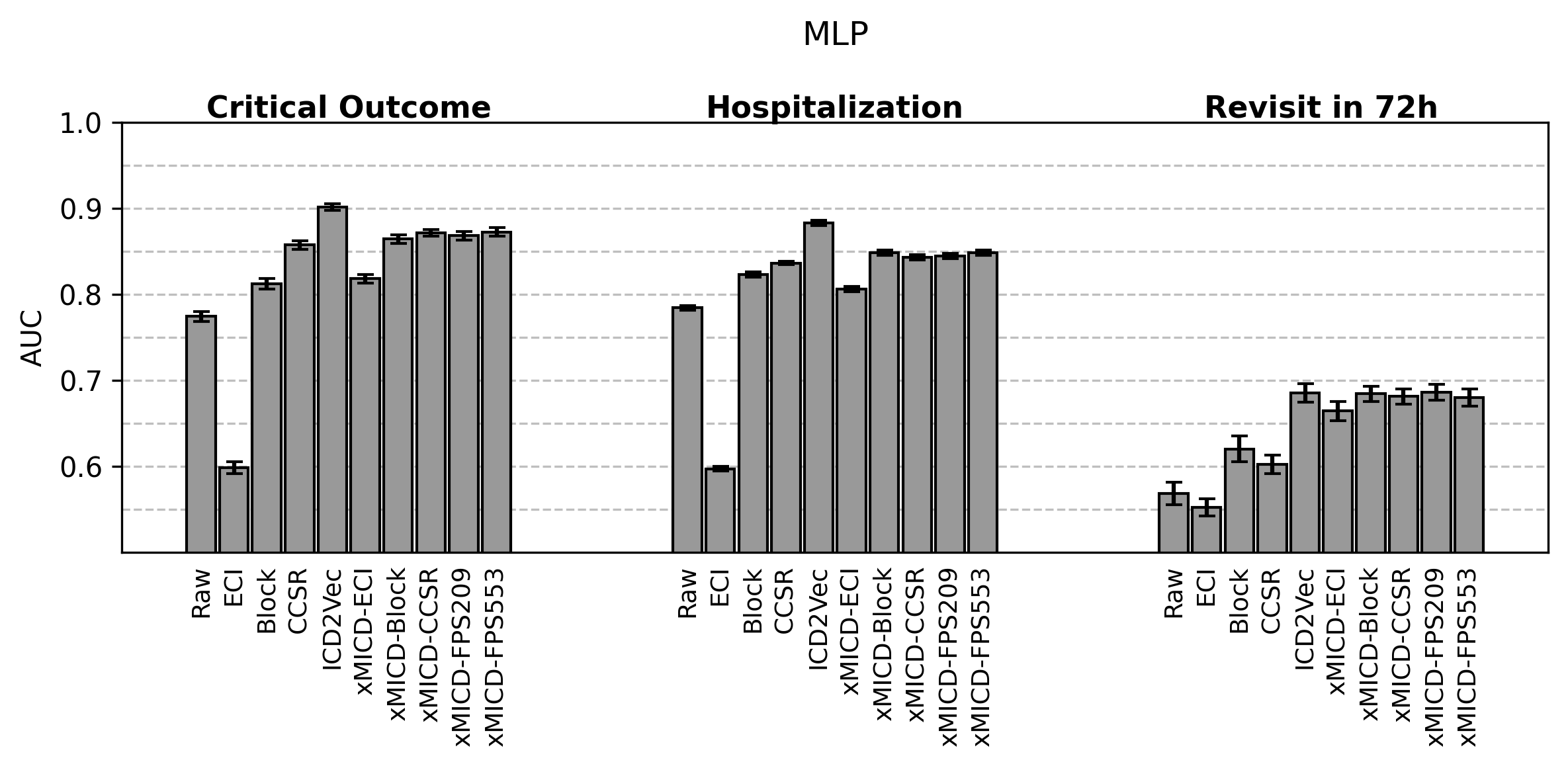}
\caption{Predictive performance of different ICD representations on the MIMIC-IV-ED dataset using MLP.}
\label{fig:mlp_task123}
\end{figure}

\begin{figure}[htbp]
\centering
\includegraphics[width=\linewidth]{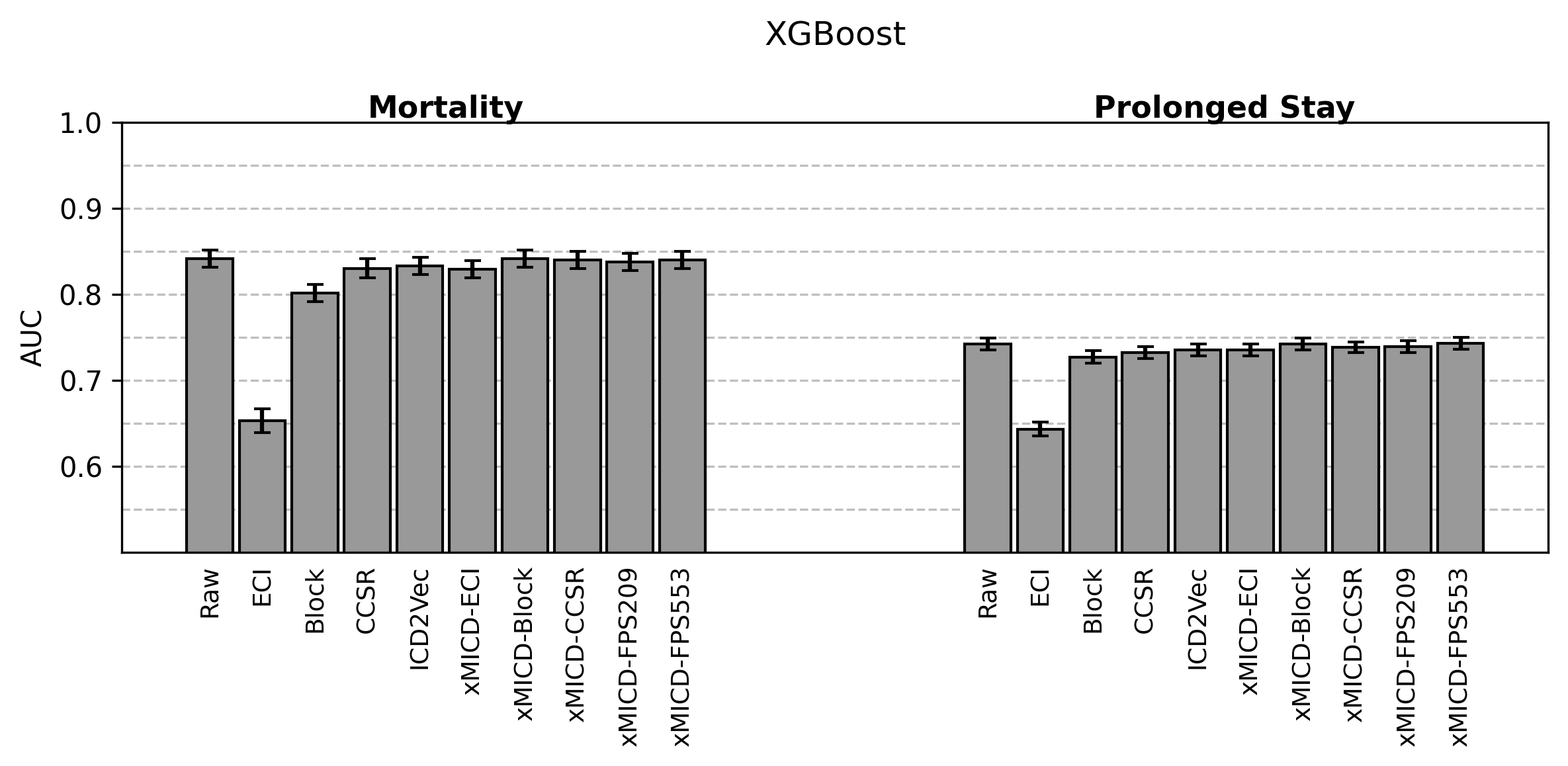}
\caption{Predictive performance of different ICD representations on the eICU-CRD dataset using XGBoost.}
\label{fig:xgboost_task45}
\end{figure}

\begin{figure}[htbp]
\centering
\includegraphics[width=\linewidth]{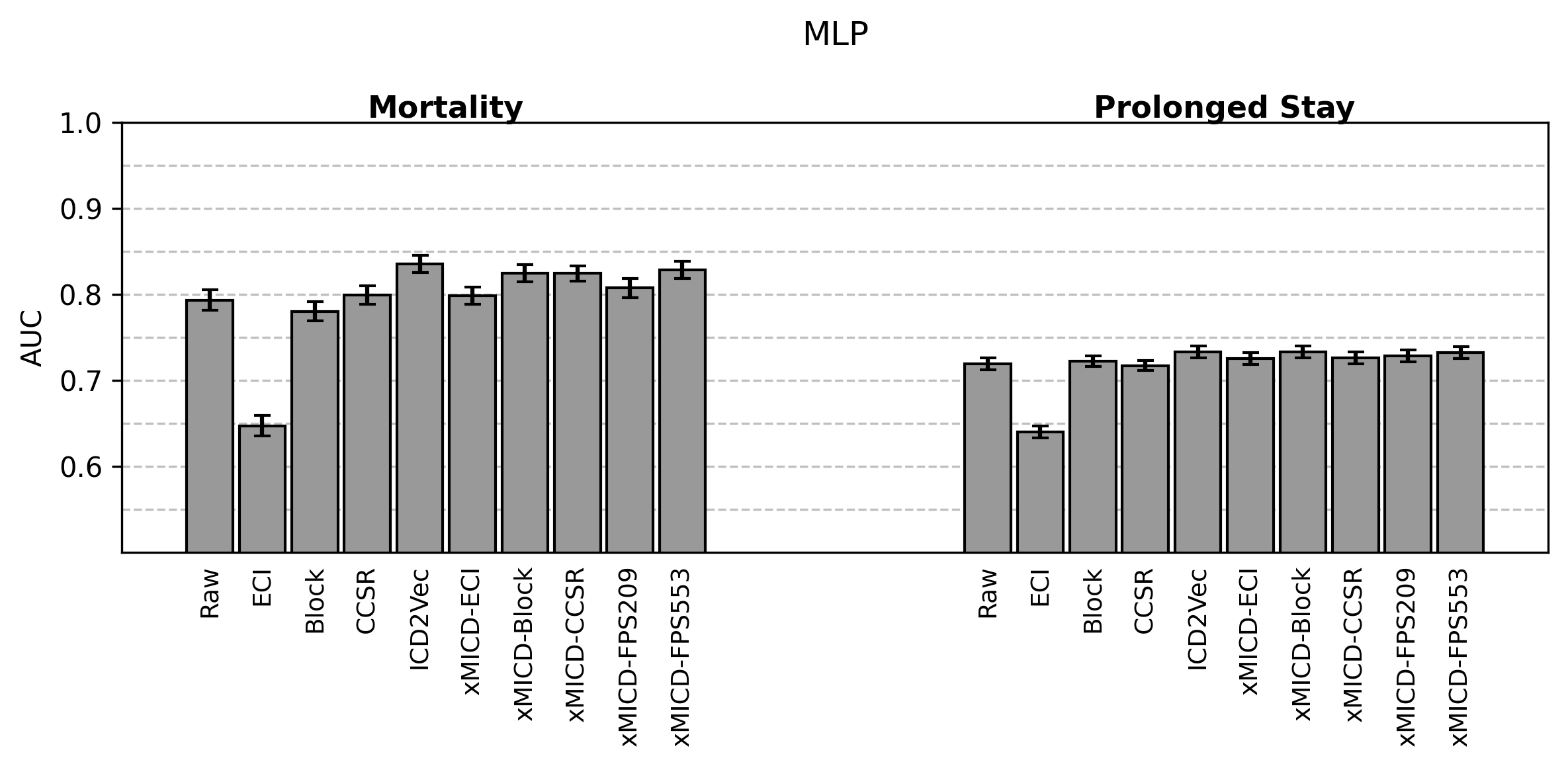}
\caption{Predictive performance of different ICD representations on the eICU-CRD dataset using MLP.}
\label{fig:mlp_task45}
\end{figure}

We evaluate xMICD constructed using cosine similarity computed from ICD2Vec embeddings with different types of anchors. We compare these variants with Raw ICD features, ICD2Vec embeddings, and several binary grouping features. Figures~\ref{fig:xgboost_task123} and \ref{fig:mlp_task123} present the results on the MIMIC-IV-ED dataset for the tasks of Critical Outcome, Hospitalization, and Revisit in 72h using XGBoost and MLP models, respectively. Figures~\ref{fig:xgboost_task45} and \ref{fig:mlp_task45} show the results on the eICU-CRD dataset for the tasks of Mortality and Prolonged Stay using the same two models. Error bars represent the 95\% confidence intervals. The complete set of results is reported in Table~\ref{tab:auc_results} and Table~\ref{tab:auc_results_eicu} in the Appendix.

Overall, we observe that binary grouping representations, which reduce dimensionality by assigning codes to discrete clinical groups, generally lead to lower predictive performance compared to using Raw ICD features. In contrast, embedding-based representations such as ICD2Vec consistently improve predictive performance across most tasks and models, suggesting that continuous representations capture richer relationships between diagnoses than binary indicators.

Among the different xMICD variants constructed using different anchor sets, most configurations yield very similar predictive performance. In particular, the difference between clinically defined anchors (such as ICD blocks or CCSR categories) and data-driven anchor (FPS) appears to be relatively small when the dimensionality is comparable. The main exception is xMICD based on ECI anchors. Because ECI contains only 31 groups, the resulting representation has a much lower dimensionality and therefore captures less diagnostic information, leading to substantially lower predictive performance than other xMICD variants.

Finally, we find that xMICD representations built on ICD2Vec embeddings, except for the ECI-based variant, achieve predictive performance comparable to ICD2Vec itself. For XGBoost models, xMICD occasionally slightly exceeds ICD2Vec, whereas for MLP models ICD2Vec typically remains marginally better. However, the differences between the two representations are generally small, indicating that xMICD preserves most of the predictive power of ICD2Vec while providing a more structured and interpretable representation. These patterns hold consistently across nearly all tasks and models considered in our experiments.

For the eICU-CRD dataset, it can be observed that Raw ICD codes perform comparably to both ICD2Vec and xMICD. This is largely due to the fact that the dataset contains only 811 unique codes, which is even fewer than the dimensionality of the ICD2Vec embeddings. As a result, dimensionality reduction provides limited benefit in this setting. In contrast, in the MIMIC-IV-ED dataset the number of distinct codes reaches the order of tens of thousands, making dimensionality reduction substantially more important.

This contrast suggests that xMICD is particularly useful in settings with a large number of unique ICD codes, such as MIMIC-IV-ED, where dimensionality reduction becomes important for improving model performance. At the same time, the results show that xMICD can preserve predictive performance at a level close to ICD2Vec while providing a more structured and interpretable representation.

\subsection{Similarity preservation}

\begin{figure}[htbp]
\centering
\includegraphics[width=0.65\linewidth]{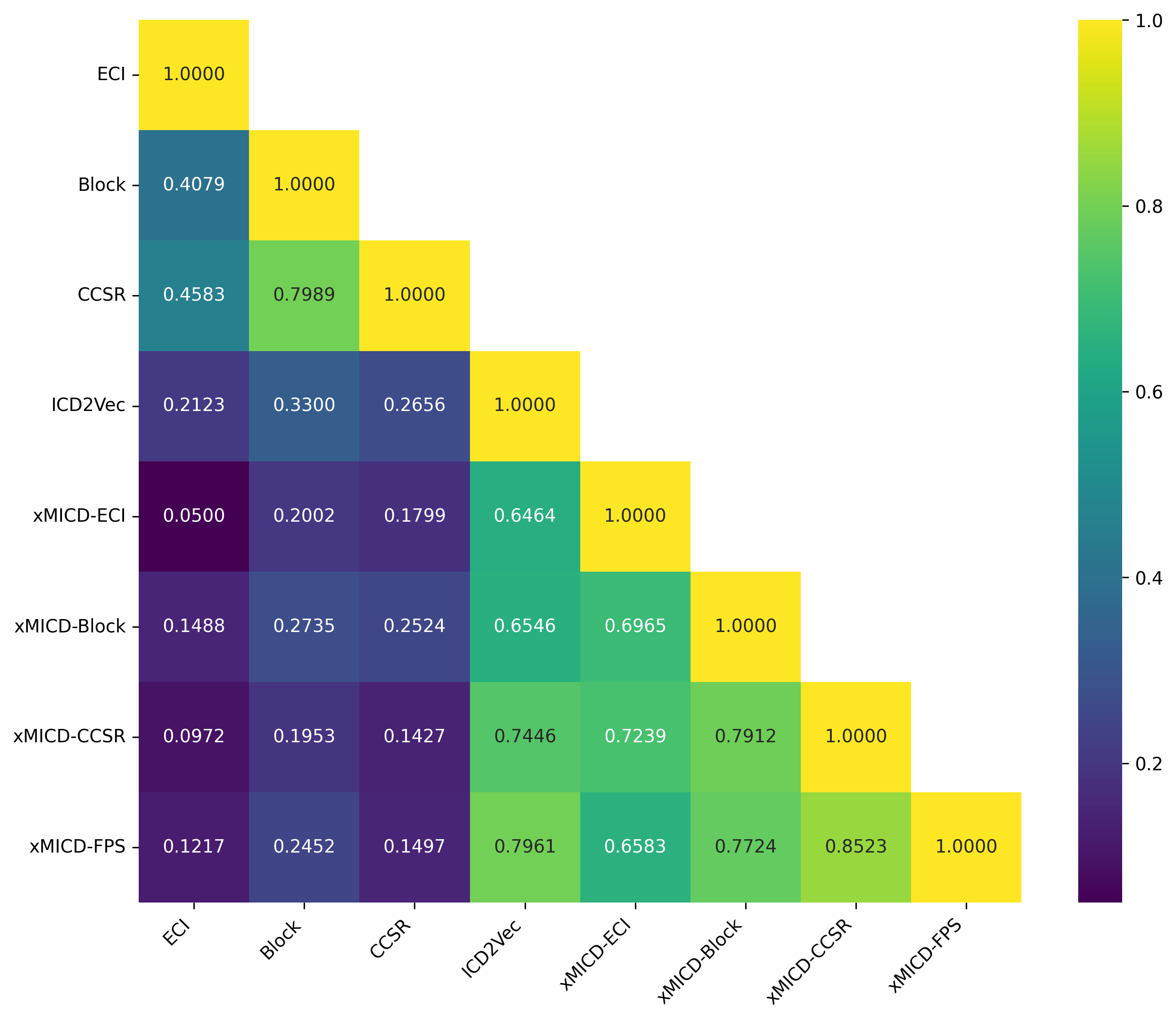}
\caption{Pairwise Spearman correlation between patient--patient similarity matrices induced by different ICD representations. Higher values indicate that two representations preserve more similar patient similarity structures.}
\label{fig:similar_heatmap}
\end{figure}

To understand why xMICD based on ICD2Vec retains strong predictive performance, we analyze how well different representations preserve the similarity structure between patients. Figure~\ref{fig:similar_heatmap} shows the pairwise Spearman correlation between patient--patient similarity matrices derived from each representation.

The results indicate that xMICD variants remain substantially closer to ICD2Vec than binary grouping-based representations. In particular, xMICD-FPS shows the highest agreement with ICD2Vec (0.7961), followed by xMICD-CCSR (0.7446), xMICD-Block (0.6546), and xMICD-ECI (0.6464). By contrast, the correlations between ICD2Vec and the binary grouping features are considerably lower: 0.2123 for ECI, 0.3300 for Block, and 0.2656 for CCSR.

These findings suggest that xMICD preserves much of the relational structure captured by ICD2Vec. Although xMICD replaces the dense latent dimensions of ICD2Vec with clinically interpretable anchor-based features, the resulting patient similarity relationships remain strongly aligned with those induced by ICD2Vec. This helps explain why xMICD can maintain predictive performance close to ICD2Vec while providing a more structured and interpretable representation.

\subsection{Interpretability}

We evaluate the interpretability of different ICD code representations according to the criteria introduced in Section~\ref{sec:criteria}: readability, understandability, meaningfulness, trackability, and simulatability.

\subsubsection{Readability, Understandability, and Meaningfulness}

Binary grouping representations are generally readable because their features correspond to clinically familiar concepts such as diagnoses, comorbidities, or disease categories. In contrast, embedding-based representations such as ICD2Vec are not directly readable because their dimensions correspond to latent numerical components without interpretable labels. The xMICD representation largely preserves readability because each dimension is anchored to a clinically meaningful diagnostic group. As a result, clinicians can associate each feature with recognizable disease categories even though the values themselves are derived from similarity calculations.

A similar contrast appears in terms of understandability. Binary grouping representations are directly understandable because their features correspond to concepts clinicians routinely use in reasoning. ICD2Vec lacks this property since its latent dimensions capture statistical relationships rather than explicit medical concepts. In xMICD, each feature represents the similarity between a patient’s diagnoses and a clinically defined group. Although these values do not correspond to a direct clinical quantity, they can be interpreted as indicators of diagnostic relatedness, allowing clinicians to reason about the representation in practice.

Meaningfulness is more nuanced. Binary grouping representations are meaningful because each feature corresponds to a specific clinical entity. ICD2Vec lacks this property since embedding dimensions do not have inherent clinical interpretation. In xMICD, the dimensions correspond to recognizable disease groups, but feature values reflect ‘diagnostic proximity’ or ‘phenotypic similarity’ rather than explicit membership. Unlike traditional binary encoding, xMICD captures the clinical neighborhood of a patient. Consequently, a patient may obtain a non-zero value for a group even if no diagnosis belongs to that group directly. This allows clinicians to identify patients who are leaning toward specific critical states. While this requires careful interpretation, the representation still preserves clinically meaningful reference concepts and provides more interpretable signals than pure embeddings.

The choice of anchors also affects interpretability. When anchors correspond to clinically defined groupings such as ICD blocks or curated disease categories, each dimension maps to a recognizable clinical concept. In contrast, anchors selected using data-driven procedures such as farthest point sampling may correspond to individual ICD codes that are less clinically intuitive. In this case the features remain interpretable as similarity to reference diagnoses, but their clinical meaning may be less immediately apparent.

\subsubsection{Trackability and Simulatability}

Regarding trackability, binary grouping representations can be traced directly to the original ICD codes through explicit mappings. ICD2Vec and xMICD are only partially trackable because they depend on embeddings derived from the code corpus. In terms of simulatability, binary grouping representations are fully simulatable because they follow deterministic rules. ICD2Vec is not simulatable without training an embedding model, whereas xMICD lies between these extremes: it depends on embeddings but uses explicitly defined similarity and aggregation rules.

Overall, the comparison reveals a clear trade-off. Binary grouping representations offer strong readability and traceability but may lose information. ICD2Vec captures rich semantic relationships but provides little interpretability. xMICD occupies an intermediate position, preserving much of the predictive signal of embedding-based representations while mapping explanations back to clinically meaningful diagnostic groups.

\subsubsection{Example of Model Explanations}

\begin{figure}[htbp]
\centering
\begin{subfigure}[b]{0.49\textwidth}
\centering
\includegraphics[width=\textwidth]{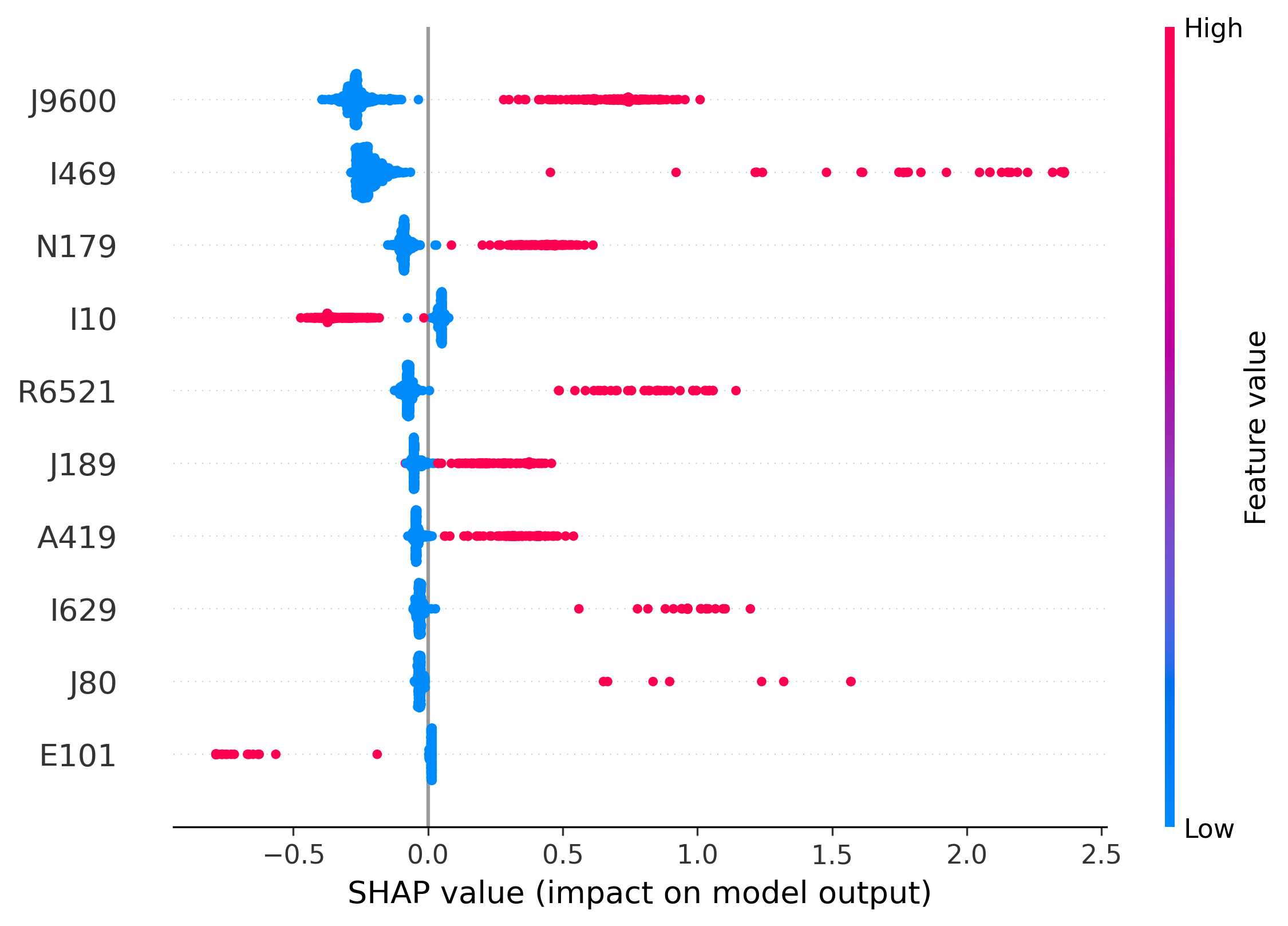}
\caption{Raw ICD}
\end{subfigure}
\hfill
\begin{subfigure}[b]{0.49\textwidth}
\centering
\includegraphics[width=\textwidth]{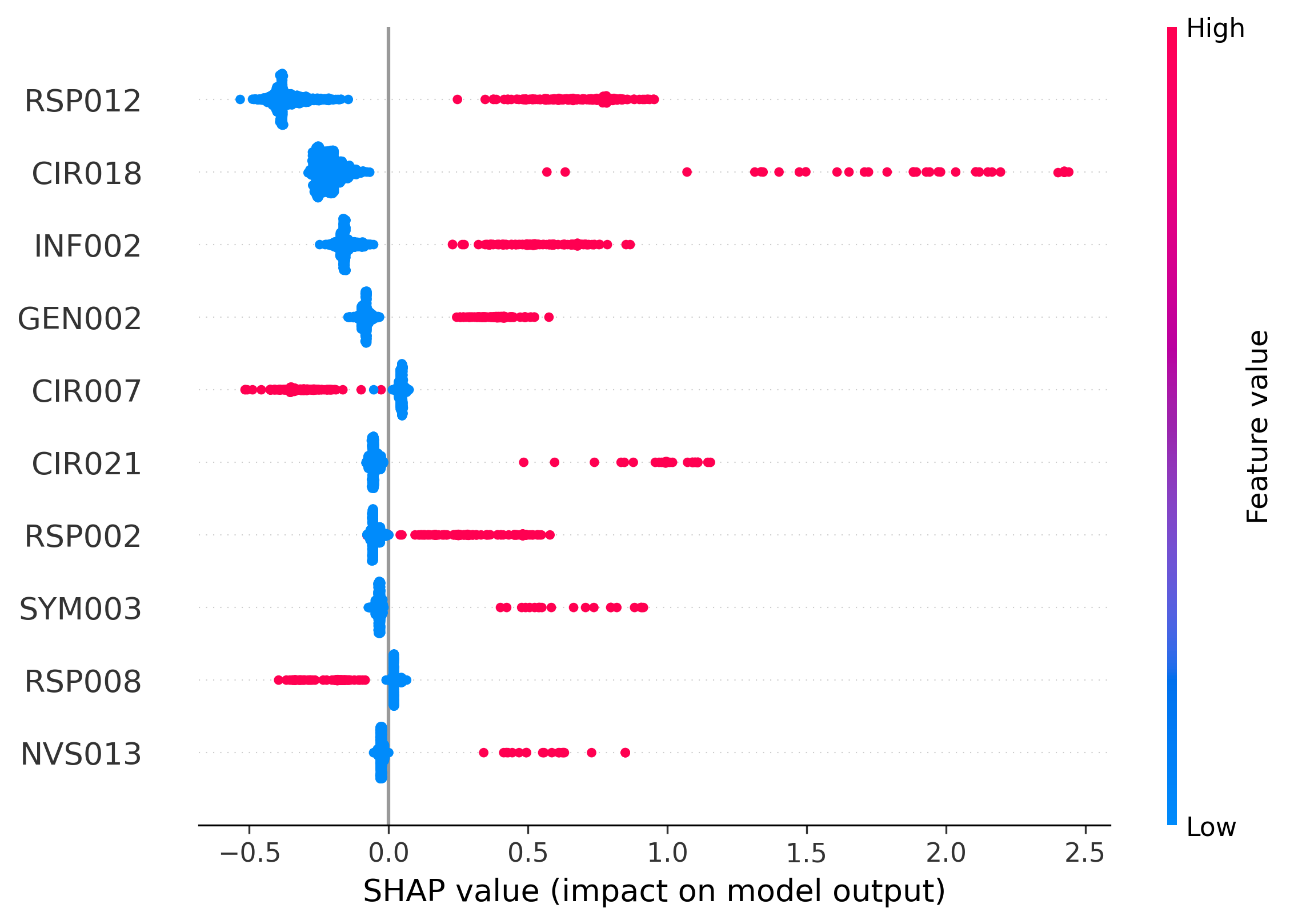}
\caption{CCSR}
\end{subfigure}

\vskip\baselineskip

\begin{subfigure}[b]{0.49\textwidth}
\centering
\includegraphics[width=\textwidth]{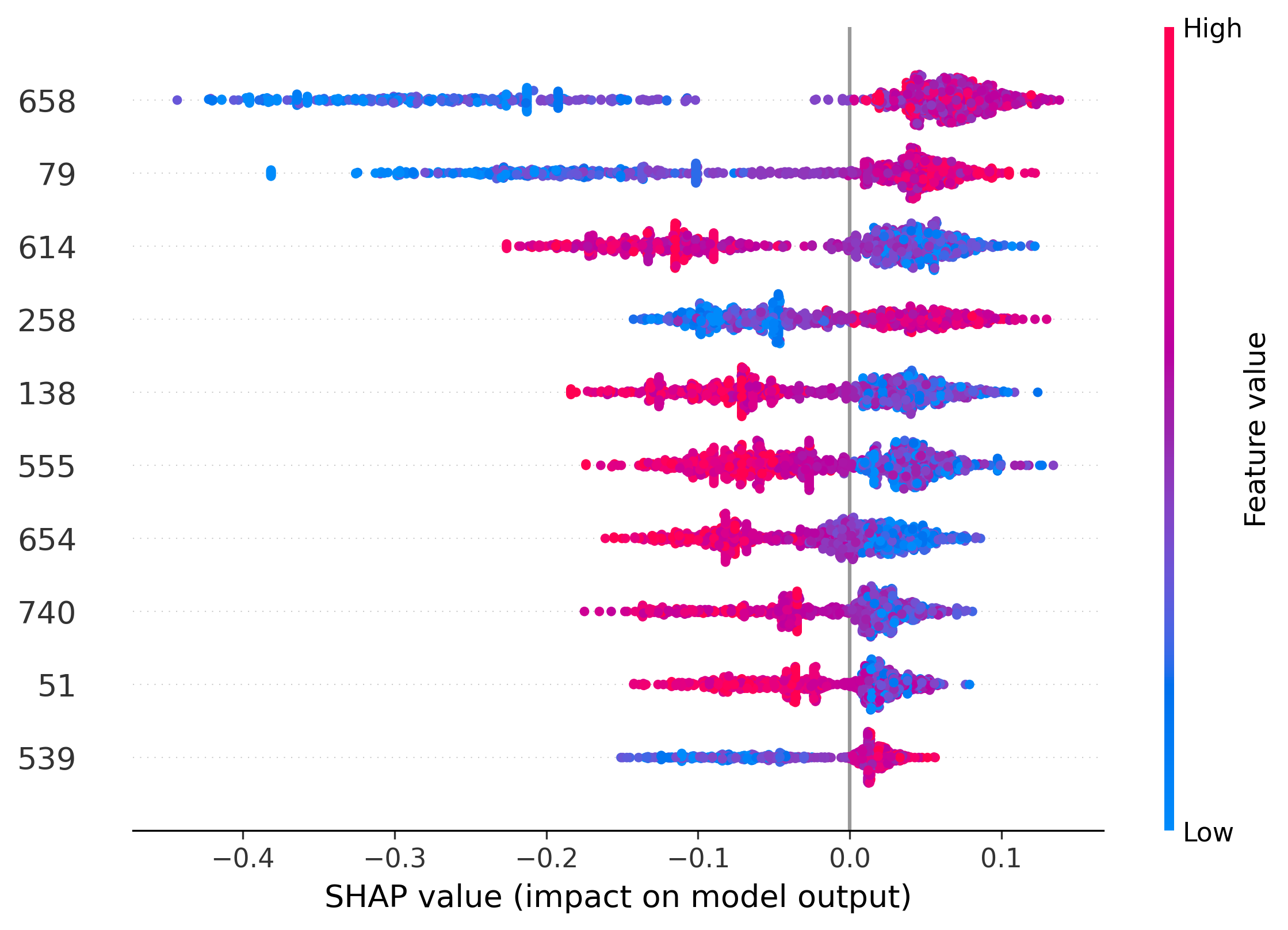}
\caption{ICD2Vec}
\end{subfigure}
\hfill
\begin{subfigure}[b]{0.49\textwidth}
\centering
\includegraphics[width=\textwidth]{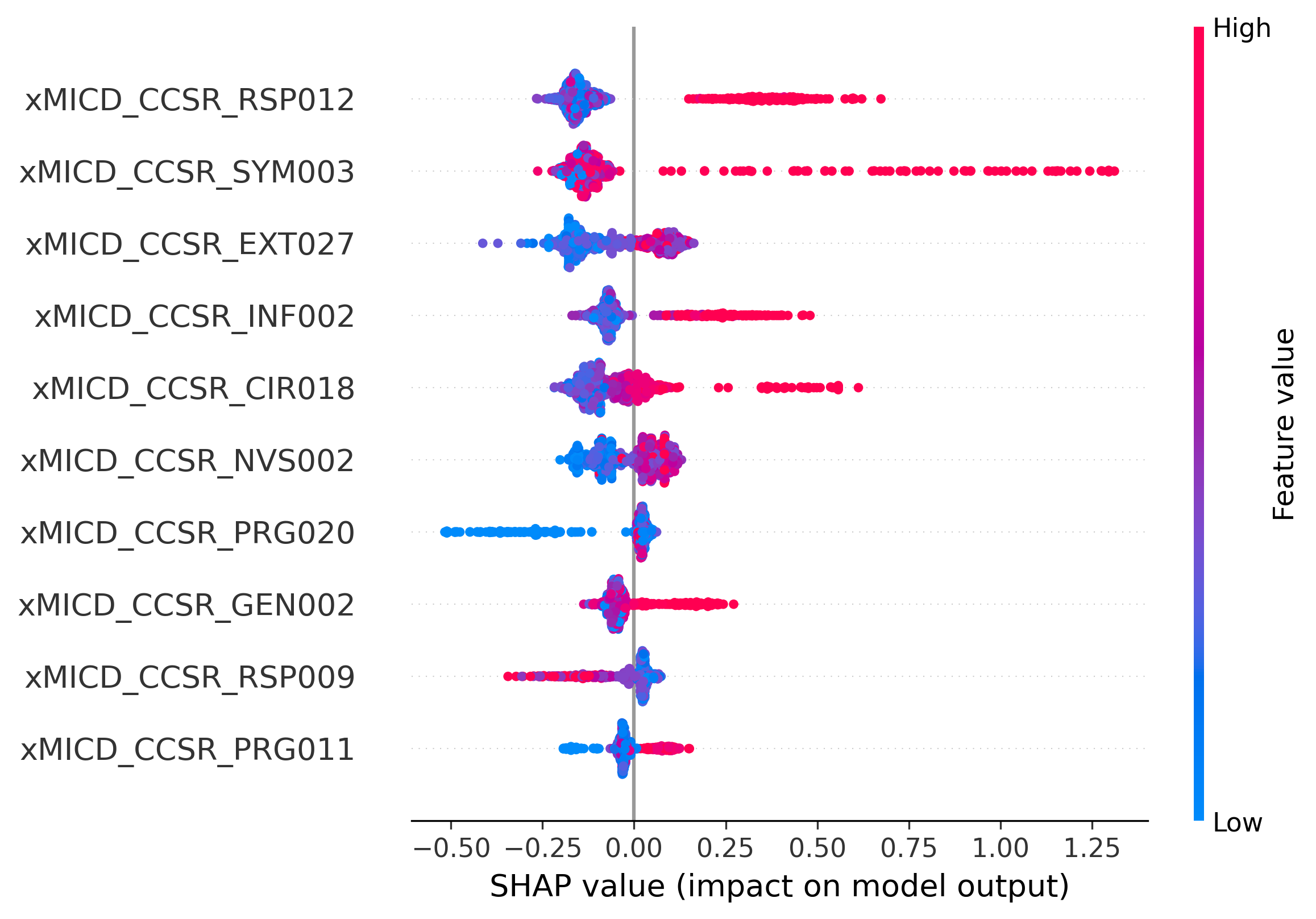}
\caption{xMICD-CCSR}
\end{subfigure}

\vskip\baselineskip

\begin{subfigure}[b]{0.49\textwidth}
\centering
\includegraphics[width=\textwidth]{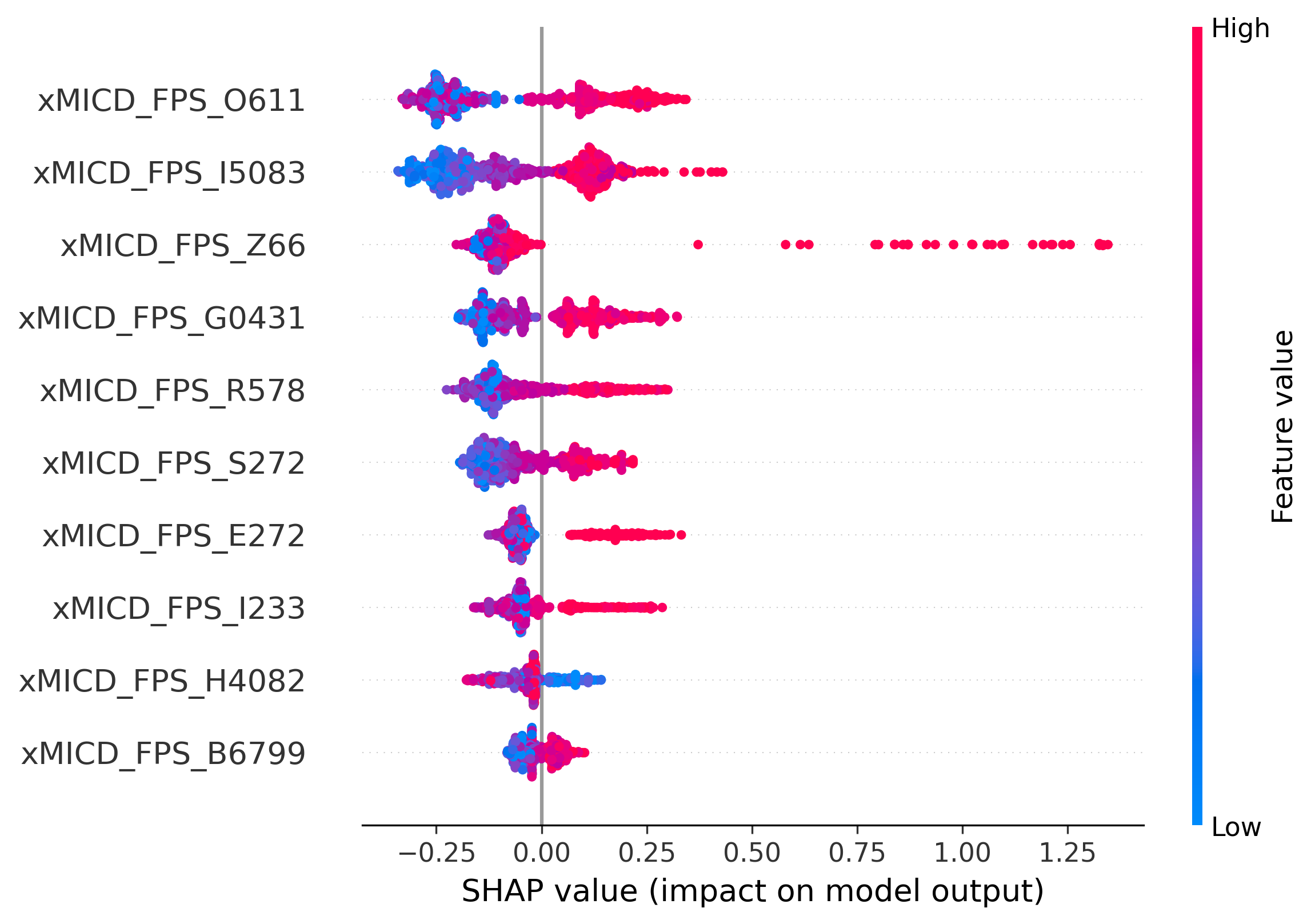}
\caption{xMICD-FPS}
\end{subfigure}

\caption{Global SHAP summary plots showing the most influential features for predicting in-ICU mortality using different ICD representations. Each point represents a patient observation. The horizontal axis shows the SHAP value indicating the impact of a feature on the prediction, while color represents the feature value.}
\label{fig:all_shap_plots}
\end{figure}

To examine interpretability empirically, we analyze global SHAP (Shapley Additive exPlanations) values aggregated across the dataset. Figure~\ref{fig:all_shap_plots} shows SHAP summary plots for predicting in-ICU mortality on the eICU-CRD dataset using an XGBoost model. Each plot displays the most influential features across the dataset, where the horizontal axis represents the SHAP value and color indicates the feature value.

Using the Raw ICD representation, the model highlights individual diagnosis codes as the most influential predictors. Severe conditions such as respiratory failure (J96.00), cardiac arrest (I46.9), and acute kidney failure (N17.9) show strong positive contributions, reflecting clinically intuitive drivers of mortality risk in the ICU. However, explanations based on raw codes are highly fragmented because each diagnosis appears as an independent sparse feature.

When ICD codes are aggregated into clinically defined groups, as in the CCSR representation, explanations become more structured. Influential predictors correspond to broader disease categories such as respiratory disorders, infections, and circulatory conditions, providing a clearer clinical overview. By contrast, ICD2Vec produces explanations based on embedding dimensions rather than medical concepts. Although these dimensions capture meaningful statistical relationships, they lack direct clinical interpretation.

The xMICD representations provide an intermediate form of explanation by combining embedding-based similarity with clinically meaningful anchors. In the xMICD-CCSR representation, each feature corresponds to the similarity between a patient’s diagnoses and a specific disease group, allowing explanations to remain clinically interpretable while still reflecting relationships learned in the embedding space.

Two observations further illustrate this behavior. First, the ranking of important disease groups remains broadly consistent between CCSR and xMICD-CCSR. Groups such as respiratory failure (RSP012), infections and sepsis (INF002), and renal failure (GEN002) appear among the most influential predictors in both representations, indicating that xMICD preserves the core clinical signals present in the data. At the same time, some groups change their relative importance because xMICD distributes similarity across related disease groups rather than using binary membership. This suggests that xMICD captures broader physiological states rather than relying solely on discrete diagnosis labels.

Second, the comparison between Raw ICD and xMICD-FPS reveals how embedding-based similarity can expose latent clinical relationships. In the Raw ICD representation, the code Z66 (Do Not Resuscitate) does not appear among the most influential predictors, likely because it is rare or inconsistently recorded. Instead, the model relies on diagnoses such as cardiac arrest, respiratory failure, or severe infection to infer critical patient status. In contrast, the ability of xMICD to link Z66 with critical anchors demonstrates its capacity for phenotype discovery. In clinical practice, a Do Not Resuscitate order is rarely an isolated event but rather a proxy for multi-organ failure or terminal illness. xMICD’s interpretability allows the model to surface these latent clinical contexts which are often lost in traditional sparse representations. This does not imply that the patient explicitly has a Z66 diagnosis, but rather that the patient’s diagnoses occupy a similar region of the embedding space associated with terminal or critical conditions.

\section{Discussion}

The results from experiments on predictive performance and interpretability analyses of different ICD code representations are summarized in Table~\ref{tab:auc_inter}. In addition to predictive performance, the table compares dimensionality and several interpretability-related criteria. Raw ICD codes and grouping-based vectorization methods exhibit high interpretability but relatively limited predictive power. In contrast, deep learning–based embedding methods such as ICD2Vec achieve strong predictive performance but are difficult to interpret, suggesting that high interpretability often comes at the cost of predictive performance. This observation aligns closely with findings reported in prior literature discussed in Section~\ref{sec:relate}.

\begin{table}[htbp]
\centering
\footnotesize 
\caption{Comparison of ICD code representations in terms of dimensionality, predictive power, and five interpretability-related properties: readability, understandability, meaningfulness, trackability, and simulatability. Symbols indicate the extent to which each property is present: \ding{108} means present, \ding{119} means partially present, and -- means absent or not present. Symbols provide a qualitative high-level summary based on the findings of this study.}
\label{tab:auc_inter}
\begin{tabular}{lccccc}
\toprule
 & \textbf{Raw} & \textbf{ECI} & \textbf{Block/CCSR} & \textbf{ICD2Vec} & \textbf{xMICD} \\
\midrule
Low-Dimension   &   --   &   \ding{108}    &  \ding{108}  &    \ding{119}    &   \ding{108}    \\
Predictive   &   \ding{119}   &   --    &  \ding{119}  &    \ding{108}    &   \ding{108}    \\
\midrule
Readable      &   \ding{108}   &   \ding{108}    &  \ding{108}  &    --    &   \ding{108}    \\
Understandable   &   \ding{108}   &   \ding{108}    &  \ding{108}  &    --    &   \ding{108}    \\
Meaningful  &   \ding{108}   &   \ding{108}    &  \ding{108}  &    --    &   \ding{119}    \\
Trackable      &   \ding{108}   &   \ding{108}    &  \ding{108}  &    \ding{119}    &   \ding{119}    \\
Simulatable      &   \ding{108}   &   \ding{108}    &  \ding{108}  &    --    &   \ding{119}    \\
\bottomrule
\end{tabular}
\end{table}

However, we demonstrate that xMICD, except for the ECI-based variant, achieves predictive performance comparable to ICD2Vec across most prediction tasks. This relatively small performance gap supports the central hypothesis of this work: high interpretability does not necessarily require sacrificing predictive power.

One reason for this comparable performance lies in semantic information preservation. xMICD is built upon the semantic structure captured by ICD2Vec embeddings. By using these embeddings to compute similarity between ICD codes and clinically meaningful anchors, xMICD incorporates latent relationships learned by the embedding model while organizing them into an interpretable feature space. In this sense, xMICD can be viewed as a structured projection of ICD2Vec representations onto clinically interpretable disease groups.

Furthermore, the strong performance of xMICD when used with XGBoost highlights a practical advantage. Tree-based models are well suited to capturing nonlinear interactions and handling sparse or heterogeneous clinical features. The xMICD representation provides structured similarity signals across disease groups, allowing these models to leverage both clinical grouping information and embedding-based relationships. As observed in several prediction tasks, including Hospitalization and Revisit within 72 hours, XGBoost models using xMICD features often achieve the best or near-best predictive performance.

xMICD also has limitations that should be acknowledged. Because the representation depends on pre-trained embeddings such as ICD2Vec, it cannot be fully simulatable from the raw ICD codes alone. This reflects a trade-off between interpretability and representational richness. In addition, although xMICD anchors its features to clinically meaningful constructs such as ICD blocks or CCSR groups, the similarity-based representation may introduce ambiguity in terms of meaningfulness. A high value in a particular dimension does not necessarily indicate the direct presence of diseases within that group, but rather reflects semantic similarity between the patient’s ICD codes and the anchor diseases. 

Without careful interpretation, this may lead to overly literal or incorrect clinical conclusions. Therefore, when applying xMICD in practice, the resulting features and explanations should be interpreted with caution and with awareness of the underlying similarity mechanism. This is crucial for building trust in AI-assisted systems, as it allows practitioners to verify the model’s reasoning against their own clinical intuition. Nevertheless, because the anchors correspond to medically meaningful disease groupings, the resulting representation remains substantially more interpretable than purely embedding-based approaches.

\section{Conclusion}

In this work, we proposed xMICD, an explainable representation of multiple ICD codes designed to balance predictive performance with clinical interpretability. xMICD is a flexible framework built upon two core components: a pre-trained ICD embedding model (e.g., ICD2Vec) and a set of clinically defined groupings (e.g., Blocks, CCSR). By combining these elements through a relative assignment mechanism, xMICD represents a patient’s diagnoses as a low-dimensional vector that captures semantic relationships among co-occurring ICD codes while remaining interpretable to clinicians.

Experiments on large-scale EHR datasets demonstrate that xMICD achieves predictive performance comparable to embedding-based representations and other state-of-the-art approaches. At the same time, the representation maintains a clear connection to clinically meaningful diagnostic groups, enabling more interpretable explanations than purely embedding-based features.

Overall, xMICD provides a practical way to integrate the semantic power of ICD embeddings into clinically interpretable feature spaces. This approach contributes toward the development of more transparent and trustworthy machine learning models for AI-assisted clinical decision support.

\section*{Acknowledgment}
This research project has been supported by Mahidol University (Fundamental Fund: fiscal year 2026 by National Science Research and Innovation Fund (NSRF), grant number FF-027/2569).




\bibliographystyle{elsarticle-harv}
\bibliography{Citations}

\section*{Appendix}

Tables~\ref{tab:auc_results} and \ref{tab:auc_results_eicu} summarize the predictive performance of different ICD representations. Table~\ref{tab:auc_results} reports the main results on the MIMIC-IV-ED dataset, including Raw ICD, binary grouping features, ICD2Vec, and xMICD variants. Table~\ref{tab:auc_results_eicu} presents the corresponding evaluation on the eICU-CRD dataset. All results are reported as mean AUC with the half-width of the 95\% bootstrap confidence interval based on 200 resamples of the test set.

\begin{table}[htbp]
\footnotesize
\centering
\caption{Predictive performance (AUC) of different ICD representations on the MIMIC-IV-ED dataset using XGBoost and multilayer perceptron (MLP) models. Features are constructed from diagnoses in the current emergency department visit and inpatient diagnoses from the preceding five years.}
\label{tab:auc_results}
\begin{tabular}{l c c c c c c c}
\toprule
 &  & \multicolumn{2}{c}{\textbf{Critical Outcome}} 
    & \multicolumn{2}{c}{\textbf{Hospitalization}} 
    & \multicolumn{2}{c}{\textbf{Revisit in 72h}} \\
\cmidrule(l){3-4} \cmidrule(l){5-6} \cmidrule(l){7-8}
\textbf{} & \textbf{Dim.} & \textbf{XGBoost} & \textbf{MLP} & \textbf{XGBoost} & \textbf{MLP} & \textbf{XGBoost} & \textbf{MLP} \\
\midrule

Raw ICD & 25383
& \cell{0.861}{±0.005}
& \cell{0.774}{±0.006}
& \cell{0.806}{±0.004}
& \cell{0.784}{±0.003}
& \cell{0.692}{±0.009}
& \cell{0.568}{±0.013} \\

\midrule

ECI & 62
& \cell{0.595}{±0.006}
& \cell{0.598}{±0.007}
& \cell{0.599}{±0.003}
& \cell{0.597}{±0.003}
& \cell{0.554}{±0.010}
& \cell{0.552}{±0.010} \\

Block & 418
& \cell{0.839}{±0.005}
& \cell{0.812}{±0.006}
& \cell{0.804}{±0.003}
& \cell{0.823}{±0.003}
& \cell{0.687}{±0.010}
& \cell{0.620}{±0.015} \\

CCSR & 1106
& \cell{0.855}{±0.004}
& \cell{0.857}{±0.005}
& \cell{0.809}{±0.002}
& \cell{0.836}{±0.002}
& \cell{0.639}{±0.011}
& \cell{0.602}{±0.011} \\

\midrule

ICD2Vec Avg. & 2048
& \cell{0.893}{±0.005}
& \cell{\textbf{0.901}}{\textbf{±0.004}}
& \cell{0.864}{±0.003}
& \cell{\textbf{0.883}}{\textbf{±0.003}}
& \cell{0.697}{±0.010}
& \cell{\textbf{0.685}}{\textbf{±0.011}} \\

\midrule
xMICD-ECI & 62
& \cell{0.868}{±0.006}
& \cell{0.818}{±0.005}
& \cell{0.838}{±0.003}
& \cell{0.806}{±0.003}
& \cell{0.690}{±0.008}
& \cell{0.664}{±0.011} \\

xMICD-Block & 418
& \cell{\textbf{0.893}}{\textbf{±0.003}}
& \cell{0.864}{±0.005}
& \cell{\textbf{0.861}}{\textbf{±0.003}}
& \cell{0.848}{±0.003}
& \cell{0.702}{±0.010}
& \cell{0.684}{±0.009} \\

xMICD-CCSR & 1106
& \cell{0.893}{±0.004}
& \cell{0.871}{±0.004}
& \cell{0.860}{±0.003}
& \cell{0.843}{±0.003}
& \cell{0.701}{±0.009}
& \cell{0.681}{±0.009} \\

xMICD-FPS & 418
& \cell{0.891}{±0.004}
& \cell{0.868}{±0.005}
& \cell{0.860}{±0.003}
& \cell{0.844}{±0.003}
& \cell{\textbf{0.706}}{\textbf{±0.008}}
& \cell{0.686}{±0.009} \\

xMICD-FPS & 1106
& \cell{0.892}{±0.004}
& \cell{0.872}{±0.005}
& \cell{\textbf{0.861}}{\textbf{±0.003}}
& \cell{0.848}{±0.003}
& \cell{0.699}{±0.009}
& \cell{0.680}{±0.010} \\

\bottomrule
\end{tabular}
\end{table}
\begin{table}[htbp]
\footnotesize
\centering
\caption{Predictive performance (AUC) of different ICD representations on the eICU-CRD dataset using XGBoost and multilayer perceptron (MLP) models.}
\label{tab:auc_results_eicu}
\begin{tabular}{l c c c c c}
\toprule
 &  & \multicolumn{2}{c}{\textbf{Mortality}} & \multicolumn{2}{c}{\textbf{Prolonged Stay}} \\
\cmidrule(l){3-4} \cmidrule(l){5-6}
\textbf{} & \textbf{Dim.} & \textbf{XGBoost} & \textbf{MLP} & \textbf{XGBoost} & \textbf{MLP} \\
\midrule

Raw ICD & 811
& \cell{\textbf{0.841}}{\textbf{±0.010}}
& \cell{0.793}{±0.012}
& \cell{0.742}{±0.007}
& \cell{0.719}{±0.007} \\

\midrule

ECI & 31
& \cell{0.653}{±0.014}
& \cell{0.647}{±0.012}
& \cell{0.643}{±0.008}
& \cell{0.640}{±0.007} \\

Block & 209
& \cell{0.801}{±0.010}
& \cell{0.780}{±0.011}
& \cell{0.727}{±0.007}
& \cell{0.722}{±0.006} \\

CCSR & 553
& \cell{0.830}{±0.011}
& \cell{0.799}{±0.011}
& \cell{0.732}{±0.007}
& \cell{0.717}{±0.006} \\

\midrule

ICD2Vec Avg. & 1024
& \cell{0.833}{±0.010}
& \cell{\textbf{0.835}}{\textbf{±0.010}}
& \cell{0.735}{±0.007}
& \cell{\textbf{0.733}}{\textbf{±0.007}} \\

\midrule

xMICD-ECI & 31
& \cell{0.829}{±0.010}
& \cell{0.798}{±0.010}
& \cell{0.735}{±0.007}
& \cell{0.725}{±0.007} \\

xMICD-Block & 209
& \cell{\textbf{0.841}}{\textbf{±0.010}}
& \cell{0.824}{±0.010}
& \cell{0.742}{±0.007}
& \cell{0.733}{±0.007} \\

xMICD-CCSR & 553
& \cell{0.840}{±0.010}
& \cell{0.824}{±0.009}
& \cell{0.738}{±0.006}
& \cell{0.726}{±0.007} \\

xMICD-FPS & 209
& \cell{0.837}{±0.010}
& \cell{0.807}{±0.011}
& \cell{0.739}{±0.007}
& \cell{0.728}{±0.007} \\

xMICD-FPS & 553
& \cell{0.840}{±0.010}
& \cell{0.828}{±0.010}
& \cell{\textbf{0.743}}{\textbf{±0.007}}
& \cell{0.732}{±0.007} \\

\bottomrule
\end{tabular}
\end{table}

\end{document}